# Alternatives of Unsupervised Representations of Variables on the Latent Space

**Alex Glushkovsky**
BMO Financial Group

**Abstract**

The article addresses the application of unsupervised machine learning to represent variables on the 2D latent space by applying a variational autoencoder (beta-VAE). Representation of variables on low dimensional spaces allows for data visualization, disentanglement of variables based on underlying characteristics, finding of meaningful patterns and outliers, and supports interpretability. Five distinct methods have been introduced to represent variables on the latent space: (1) straightforward transposed, (2) univariate metadata of variables, such as variable statistics, empirical probability density and cumulative distribution functions, (3) adjacency matrices of different metrics, such as correlations, $R^2$ values, Jaccard index, cosine similarity, and mutual information, (4) gradient mappings followed by spot cross product calculation, and (5) combined. Twenty-eight approaches of variable representations by beta-VAE have been considered. The pairwise spot cross product addresses relationships of gradients of two variables along latent space axes, such as orthogonal, confounded positive, confounded negative, and everything in between. The article addresses generalized representations of variables that cover both features and labels. Dealing with categorical variables, reinforced entanglement has been introduced to represent one-hot encoded categories. The article includes three examples: (1) synthetic data with known dependencies, (2) famous MNIST example of handwritten numbers, and (3) real-world multivariate time series of Canadian financial market interest rates. As a result, unsupervised representations of interest rates on the latent space correctly disentangled rates based on their type, such as bonds, T-bills, GICs, or conventional mortgages, positioned bonds and T-bills along a single curve, and ordered rates by their terms along that curve.

## 1 Introduction

Representations of high dimensional data on a low dimensional space have been in focus for a long time. It covers two major objectives: visualization and encoding. Visualization allows for exploration of data to find dependencies, patterns, distinct segments, or outliers. All that can guide the following decisions and corresponding actions. Encoding transforms original data into low dimensional vectors and allows for dimensionality reduction while preserving some important inherent dependencies.

In the context of this paper, original data assumes to be formatted as the most popular relational data structure, where rows represent observations, while columns attribute variables, such as features and labels. Essentially, rows can be linked to the identification of records, while columns support understanding and interpretations.

It can be noted that unprecedented attention has been given to represent observations on a low dimensional space while representation of variables has limited coverage. Thus, principal component analysis (Jolliffe and Cadima, 2016; Rolinnek et al, 2019), observation clustering (Oyewole and Thopil, 2023), autoencoders (Bank *et al*, 2020), adversarial autoencoders (Makhzani *et al*, 2016), t-SNE (Van der Maaten and Geoffrey, 2008), InfoGAN (Chen *et al*, 2016), InfoVAE (Zhao *et al*, 2018), and UMAP (McInnes *et al*, 2018), just to name some methodological approaches, have been developed and utilized mostly for observations.

Concerning representations of variables, one can note univariate approaches, such as simple statistics (mean, standard deviation, etc.) or fitting of distributions. The most popular multivariate representation is variable clustering (Kojadinovic, 2004; Sanch and Lonergan, 2006; Fuchs and Wang, 2024).

The article addresses unsupervised representation of variables on 2D space by a variational autoencoder. Representation on 2D space supports visualization while the variational autoencoder allows for representations of new data on already encoded 2D space to observe changes or represent dynamic trajectories.

The article addresses variables that include both features and labels, as well as continuous numeric and categorical data types.

Unsupervised representation does not require data labelling. In case of a large number of variables, it allows for their reduction by selecting distinct variables while dropping some neighboring ones. In cases when labels are present in data, they can be part of the represented variables on the low dimensional space. It permits exploration of dependencies of labels to other variables. The latest can be used to screen for features preceding modeling against labels or to support interpretability and insight discoveries.

The beta variational autoencoder (β-VAE) has been selected as an unsupervised method of representing high dimensional data on the low dimensional latent space acknowledging the following properties: (1) capability to control posterior distribution closeness to the isotropic normal distribution, (2) ability to represent a new observation on the latent space by the pretrained encoder, and (3) generative to decode any point on the latent space as input variables (Kingma and Welling, 2014; Higgins *et al*, 2017; Dupont, 2018).

Implementation of β-VAE for representation of input variables has been introduced in (Glushkovsky, 2020b). It has "rediscovered" Madelung's rule of electron configuration in atoms and outlined the area of the rule violation.

Variables can be represented using different inputs and metrics. For example, (Glushkovsky, 2020a) article addresses representation of univariate cumulative distribution functions (CDF) on the latent space. The latent variables can be seen as metadata of encoded distribution types. It has been shown that representations of variables based on the CDF on the 2D or even 1D latent spaces provide compressed views of types of distributions and their parameters. Furthermore, it has the capability to represent dynamic trajectories indicating time related changes. However, the univariate approach essentially falls short of representing multivariate relationships between variables.

The article is comprised of four major parts:
1. Methodological approaches and challenges of representing variables by β-VAE
2. Synthetic data to study known dependencies on the latent space by applying different methodological approaches of representations
3. MNIST example of handwritten 0-9 numbers that includes categorical labels and their representation along features by applying reinforced entanglement
4. Real-world multivariate time series of Canadian financial market interest rates

## 2    Methodological Approaches and Challenges of Variable Representations

### 2.1 Applied Data Transformations and Metrics

Five distinct methods of variable representations by β-VAE have been addressed in the article (Figure 1) including twenty-eight alternatives:
- A. Straightforward transposed approach by flipping observations with variables
- B. Transposed approach based on univariate metadata of variables:
   - statistics, such as mean, standard deviation, entropy, Kolmogorov-Smirnov distance to uniform distribution, and distribution symmetry metric
   - empirical Probability Density Functions (PDF)
   - empirical Cumulative Distribution Functions (CDF). The latest has been determined based on a 26x25 grid approach (Glushkovsky, 2020a)
- C. Adjacency matrices (A) approach based on different pairwise metrics:
   - Correlation coefficients of the following three methods: Kendall, Pearson, and Spearman
   - $R^2$ values of the above three correlation methods
   - Jaccard index
   - Jaccard linear index
   - Cosine similarity (i.e., the dot product)
   - Mutual information

   Transformed matrices mentioned above:
   - Laplacian (D – A) matrices, where D is the degree matrix. It has not been applied to correlations and cosine similarity matrices that have negative values
   - |J – A|, where J is all-ones (i.e., the unit matrix). It converts similarity values to have a 'distance' meaning. It has not been applied to correlations, cosine similarity, and mutual information matrices that have negative values or values above one
- D. Gradient mappings of variables on the 2D observation latent space that have been converted to vectors and then to the magnitudes of the pairwise cross products for any given point (spot) on the observation latent space. Four aggregation methods of spot magnitudes of the cross product relationships and, consequently, four representations of variables have been considered in the article and outlined in paragraphs 2.2 - 2.4.
- E. Combination of methods above

Mutual information, Jaccard, and linear Jaccard indexes have been calculated based on empirical bi-dimensional probability densities. The latter has been introduced to emphasize linear dependencies between variables where the reference 2D distribution has been forced to be a diagonal one.

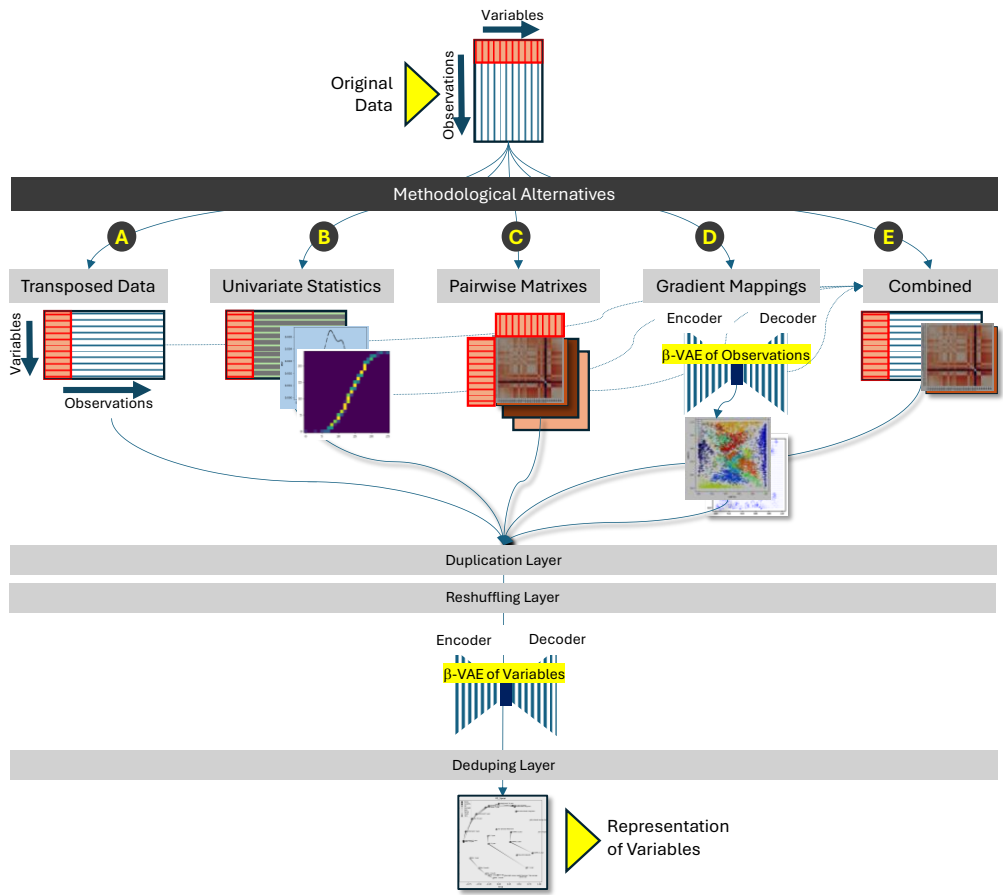

Figure 1. Diagram of the applied methodological approaches to represent variables

High-level properties of the applied methods are highlighted in Table 1.

| Flow | Method | Property | | | |
|---|---|---|---|---|---|
| | | Multivariate | Pairwise Matrix | Sign Sensitive | Gradient Based |
| A | Transposed dataframe | ✓ | | ✓ | |
| B | Transposed univariate statistics | | | | |
| | Transposed empirical probability density function | | | | |
| | Transposed empirical cumulative density function | | | | |
| C | Pairwise cosine similarity | ✓ | ✓ | ✓ | |
| | Pairwise Kendall correlation | ✓ | ✓ | ✓ | |
| | Pairwise Pearson correlation | ✓ | ✓ | ✓ | |
| | Pairwise Spearman correlation | ✓ | ✓ | ✓ | |
| | Pairwise Kendall R-sqr values | ✓ | ✓ | | |
| | Pairwise Kendall R-sqr values with Laplacian transformation | ✓ | ✓ | | |
| | Pairwise Pearson R-sqr values | ✓ | ✓ | | |
| | Pairwise Pearson R-sqr values with Laplacian transformation | ✓ | ✓ | | |
| | Pairwise Spearman R-sqr values | ✓ | ✓ | | |
| | Pairwise Spearman R-sqr values with Laplacian transformation | ✓ | ✓ | | |
| | Pairwise Jaccard index | ✓ | ✓ | | |
| | Pairwise Jaccard index with Laplacian transformation | ✓ | ✓ | | |
| | Pairwise linear Jaccard index | ✓ | ✓ | | |
| | Pairwise linear Jaccard index with Laplacian transformation | ✓ | ✓ | | |
| | Pairwise mutual information | ✓ | ✓ | | |
| | Pairwise mutual information with Laplacian transformation | ✓ | ✓ | | |
| D | Spot magnitudes of the cross product, aggregation option 1 | ✓ | ✓ | | ✓ |
| | Spot magnitudes of the cross product, aggregation option 1 with Laplacian transformation | ✓ | ✓ | | ✓ |
| | Spot magnitudes of the cross product, aggregation option 2 | ✓ | ✓ | | ✓ |
| | Spot magnitudes of the cross product, aggregation option 3 | ✓ | | | ✓ |
| | Spot magnitudes of the cross product, aggregation option 4 | ✓ | | | ✓ |

Table 1. Properties of the applied methodological approaches

The table above can be used as a guide in selecting methods based on the desired properties.

Sign sensitivity, i.e., direction of the dependency between variables, intrinsic only for the transposed data frame, cosine similarity and correlation coefficients.

As expected, there are quite strong non-linear dependencies between pairwise input matrices except for Jaccob linear indexes based on data of the synthetic example (Figure 2).

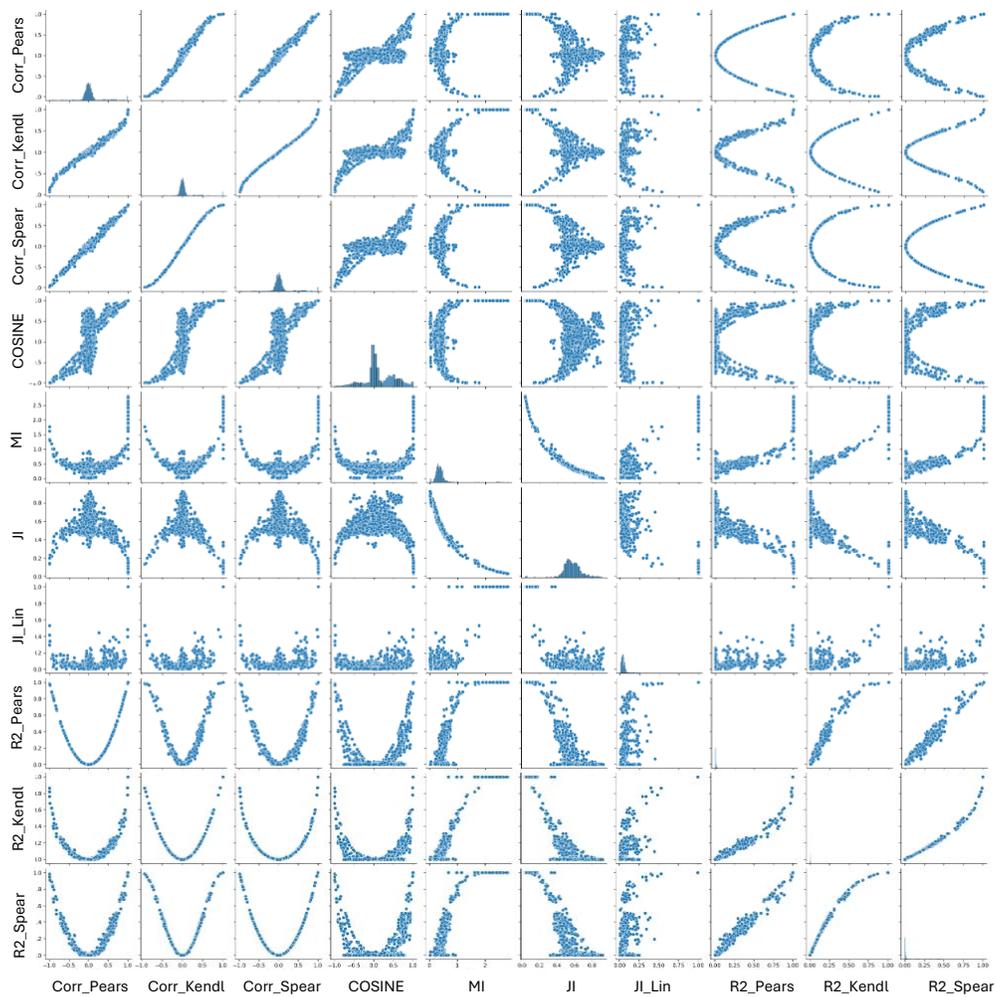

Figure 2. Dependencies between pairwise input matrices

It can be observed that the cosine similarity index, followed by Jaccard index, and mutual information, provides a wider variance then correlation coefficients or $R^2$ values for areas around zero. It is caused by types of data distributions. Thus, normally distributed cases support more monotonic relationships between the above-mentioned metrics compared to non-normally distributed data, such as uniform or exponential.

## 2.2 Challenges of Variable Representations by β-VAE

Variational autoencoding is a stochastic process, and therefore, the representations on the latent space vary from fitting to fitting. It can be attributed mostly to the isotropic property of the posterior distribution towards 2D normal distribution. Of course, initial values of weights and hyperparameters also contribute to the outcomes.

To increase robustness of the encoded results, β-VAE has been fitted three times for each method selecting the run with minimum correlation between two latent variables. The rationale for that selection is influenced by the isotropic property of the posterior distribution.

Each variable corresponds to a single column in a table of the relational database. It poses a methodological challenge of applying common machine learning approaches that requires a split between training and testing subsets. To overcome this issue, duplications of the prepared input data for variable representations and the following deduping of the encoded results have been applied (Figure 1). It should be noted that duplications have been done without adding Gaussian normally distributed random noise ~ α·N(0,1) that assumably can prevent overfitting (Guozhong An, 1996). It turned out that β-VAE could be run against the duplicated dataset without adding Gaussian noise (Glushkovsky, 2020b).

To increase robustness of the obtained results, random reshuffling of the input data has been applied after duplications.

Prepared input data for autoencoding may have sizing or balance issues concerning counts of rows and columns issues especially using transposed (A) or gradient mappings (D) methods shown in Figure 1. To mitigate these issues, a random subsampling across rows or columns has been applied.

## 2.3 Representations of Observations and Vectors of Gradients of Variables on the Latent Space

Traditional representations of observations on the latent space (Figure 3, a) support not only visualization of observations, compression of high dimensional original variables into low dimensional space, disentanglement of the underlying characteristics of the input variables, generation of new observations by decoding coordinates of any point on the latent space into a vector of original variables, but it also organizes observation points on the latent space. Even though, coordinates of the latent space (such as 2D in our case) have no standalone meaning, it provides observable mapping of variables (Figure 3, b), introduces some order along coordinates of the latent space, and, therefore, creates the possibility to estimate numerical gradients of variables $V_i$:

$$\nabla_{L_k}(V_i(L_k))$$

along latent space axes $L_{k=1:2}$.

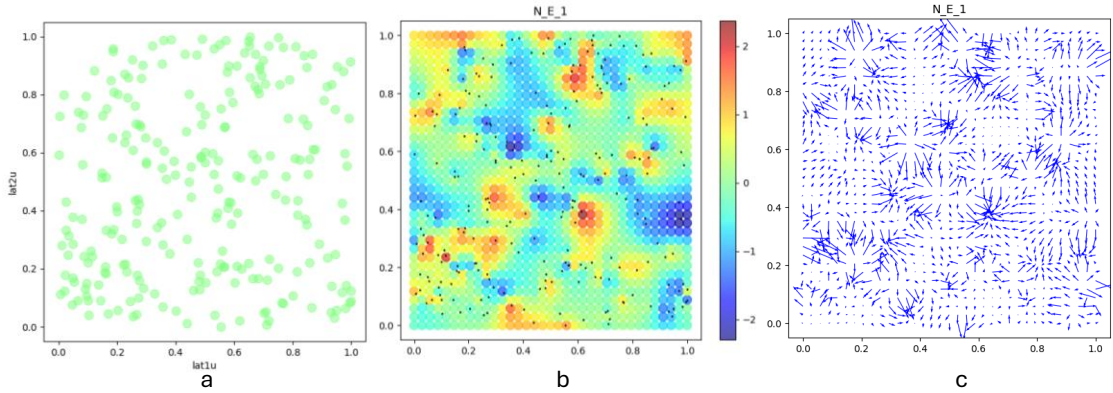

Figure 3. Examples of representation of observations on the 2D latent space:
(a) 250 observations of the synthetic data (see Section 3.1)
(b) mapping of the N_E_1 variable by applying linear interpolation on a 35x35 grid
(c) gradient vectors of N_E_1 variable

To estimate gradient vectors for each variable, the following two steps have been performed. First, linear interpolations of values of the variable applying the 35x35 grid on the 2D uniformed latent space. Second, numeric estimation of the gradients along the 2D axis. It allows for estimation of the gradient vectors for each observation (Figure 3, c). Considering vectors of any two variables for the same point on the latent space, estimation of the pairwise spot magnitude of the cross product between those variables is now possible.

## 2.4 Pairwise Spot Magnitude of Cross Product Relationships

Gradients of variables have been presented as vectors and then contrasted against each other for any given point on the latent space. It allows for estimation of spot pairwise relationships, such as cross product, based on vectors of gradients. It can be done by applying the following five steps:
1. Selection of two variables $V_i$ and $V_j$ of interest
2. Transformation of the isotropic normally distributed 2D latent space into uniformed space (0,+1; 0,+1) by applying normal CDF against the standardized latent space variables: $Lu_{1,2} = \Phi((L_{1,2} - \mu_{1,2})/\sigma_{1,2})$, where $\Phi$ – is CDF of the standard normal distribution
3. Interpolation of selected variable values on the observation latent space. It turns out that a simple linear interpolation on a regular grid applied to the latent space, such as 35x35 points, is the first choice that prevents overfitting

4. Numeric estimation of gradients *xV* and *yV* along the latent space X- and Y- axes
5. Calculation of the magnitude of the spot cross product:
$$CP_{i,j}^n = |xV_i^n * yV_j^n - yV_i^n * xV_j^n|,$$
where spot *n* can either be the point on the defined regular grid, or the nearest observation to that point.
Or:
$$CP_{ij}^n = r_i^n * r_j^n * |\sin(\alpha_i^n - \alpha_j^n)|,$$
where $r_{i,j}^n = \sqrt{(xV_{i,j}^n)^2 + (yV_{i,j}^n)^2}$ – are lengths of the gradient vectors, and $(\alpha_i^n - \alpha_j^n)$ is an angle between vectors.

It should be noted that step 3 can be omitted, however, performing interpolations on a regular grid allows for a simple numeric estimation of gradients having equal steps along the latent space axes.

Pairwise spot relationships between variables can be estimated for any observation and, therefore, possess "*curse of dimensionality*" of $O(N*K^2)$ size (Table 2).

| | | Original variables | | | Latent space | | Gradients | | | | | Magnitude of the cross product | |
|---|---|---|---|---|---|---|---|---|---|---|---|---|---|
| Obs | N_E_1 | ... | N_S_1n5 | ... | Lu1 | Lu2 | x_N_E_1 | y_N_E_1 | ... | x_N_S_1n5 | y_N_S_1n5 | ... | N_E_1 x N_S_1n5 | ... |
| 0 | -0.506 | ... | 0.660 | ... | 0.949 | 0.058 | 0.339 | -0.177 | ... | -0.019 | 0.306 | ... | 0.100 | ... |
| 1 | -0.298 | ... | 1.088 | ... | 0.943 | 0.031 | 0.339 | -0.177 | ... | -0.019 | 0.306 | ... | 0.100 | ... |
| 2 | -0.373 | ... | 0.535 | ... | 0.298 | 0.562 | 0.138 | 0.032 | ... | 0.111 | -0.399 | ... | -0.059 | ... |
| ... | ... | ... | ... | ... | ... | ... | ... | ... | ... | ... | ... | ... | ... | ... |
| 71 | 1.937 | ... | -2.004 | ... | 0.238 | 0.859 | 0.239 | 0.191 | ... | -0.249 | -0.279 | ... | -0.019 | ... |
| ... | ... | ... | ... | ... | ... | ... | ... | ... | ... | ... | ... | ... | ... | ... |
| 249 | 0.168 | ... | -0.231 | ... | 0.356 | 0.283 | -0.611 | 0.233 | ... | 0.576 | -0.206 | ... | -0.008 | ... |

(N - observations on the left; K - variables, 2D - latent, 2K - gradients, K² - pairwise at the bottom)

Table 2. Examples of the pairwise spot magnitude calculations of the cross products

The pairwise spot cross product addresses relationships between the changes of two variables, such as orthogonal, confounded positive, confounded negative, and everything in between (see legend in Figure 4).

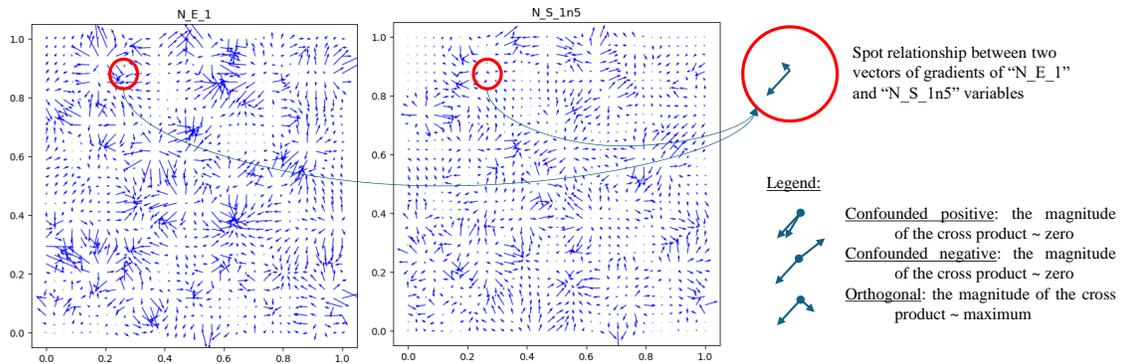

Figure 4. Example of the pairwise spot relationship between two variables (N_E_1 and N_S_1n5) based on their vectors of gradients on the observation 2D latent space

Independent spot variances of two variables assume the following conditions to be present on the same spot on the latent space: (1) significant changes across the latent space, and (2) their pairwise direction to be close to orthogonal. The first condition means that variances exist and can be considered, and the second condition indicates that changes of both variables are independently interpretable, i.e., not confounded. Satisfaction of those two conditions leads to higher pairwise spot magnitudes of the cross products. The concept of the pairwise spot magnitude of the cross product on the latent space is contrasted to correlation, $R^2$, or the cosine similarity index by the following properties: (1) it is a spot metric, (2) it is based on a gradient field that maps changes across the observation latent space, and (3) it emphasises orthogonality.

## 2.5 Applied Aggregations of Pairwise Spot Magnitudes of the Cross Products

The magnitude of cross products $CP_{ij}^n$ on 2D observation latent space have the following indexation: *i, j* - indexes of two involved variables in a pairwise relationship, and *n* – is the observation index. Having $N*K^2$ structure of $CP_{ij}^n$ data, aggregation is required to represent variables on the latent space. The following four aggregation options have been considered:

1. Arrangement of the pairwise matrix where values of the matrix equal averages of the magnitudes of the cross products across all observations $n = [1:N]$
$$\overline{CP_{ij}} = \frac{1}{N}\sum_{n=1}^{N} CP_{ij}^n$$
   The resulted pairwise matrix $\overline{CP_{ij}}$ has been used as an input for autoencoding.

2. Computation of average values of x- and y- gradients for each variable $i = [1:K]$ across all observations $n$, that produces an average gradient vector across all observations for each variable:
$$\overline{xV_i} = \frac{1}{N}\sum_{n=1}^{N} xV_i^n,$$
$$\overline{yV_i} = \frac{1}{N}\sum_{n=1}^{N} yV_i^n,$$
   and then calculation of the pairwise magnitudes of the cross products:
$$mCP_{ij} = \overline{xV_i} * \overline{yV_j} - \overline{yV_i} * \overline{xV_j}$$
   The resulted pairwise matrix $mCP_{ij}$ has been used as an input for autoencoding.

3. Reshaping magnitudes of the cross products $CP_{ij}^n$ data as an array having variables $i$ along rows and $(j,n)$ variables and observations concatenated as columns. In that case, the number of columns is significantly ($N$-times) higher than the number of rows ($K$). To balance data for autoencoding, the randomized subsampling of columns has been applied. The obtained dataset has been used as an input to train β-VAE. This approach is subsampled transposing.

4. Reshaping magnitudes of the cross products $CP_{ij}^n$ data as an array having observations and variables $(i,n)$ along rows and variables $j$ along columns. To make β-VAE training more practical, the randomized subsampling of rows has been applied. After fitting β-VAE using the reshaped and subsampled data as an input and achieving representation of the observations and variables $(i,n)$ on the latent space, the average values of the latent variables have been calculated per each variable $i$ across observations $n$.

It should be noted that aggregations of the first three options have been done prior to fitting β-VAE, while aggregation of the last option has been performed after fitting the autoencoder.

## 3  Synthetic Example

### 3.1  Data

Synthetic data has been generated to provide controllable dependencies between variables. Overall, 65 variables with 250 observations each have been created in the dataset, including 13 independent variables of four underlying distribution types and 52 dependant variables of different types of relationships. Table 3 outlines the applied factors and their levels.

| Factors | | |
| --- | --- | --- |
| Underlying distribution types | Dependencies between variables | Sign of dependence |
| Bernoulli | N/A | N/A |
| Exponential | Additive between two variables | Negative |
| Normal | Interactive between two or three variables | Positive |
| Uniform | Circular | |
| | Linear | |
| | Quadratic | |

Table 3. Synthetic data: factors and their levels

The following notation has been applied to identify the variables:
- the first character indicates the distribution type: B - Bernoulli, E - Exponential, N - Normal, or U - Uniform
- the third character underlines the type of dependencies, if any:
    - A – additive between two independent parent variables
    - E – explanatory variable (i.e., independent variable)
    - I – interactive between two or three independent parent variables (i.e., the product)
    - C, Q, or S – univariate circular, quadratic, or single linear relationships, correspondingly

- the numeric characters between the fifth and seventh positions represent identification of either the independent variable or the parent variables
- In case of a single linear relationship, n or p characters represent a sign of the dependency (i.e., negatively or positively correlated, correspondingly), while the following number 3, 5, 7, or 9 equals the strength of the linear dependency versus the normally distributed noise, i.e., 30%/70%, 50%/50%, 70%/30%, or 90%/10%, respectively.

Dependant variables have been subjected to addtional small normally distributed noise.

To simplify interpretations of the obtained results, only first order dependencies have been introduced between variables.

Here are some examples of the variable naming and their generative codes:

N_E_2 – means a normally distributed independent variable where id equals 2:
norm.rvs(size=250, loc=0, scale=1)

U_C_1 – means a variable that forms a circular relationship with an independent uniform variable where id equals 1:
((2*bernoulli.rvs(size=250, p=0.5)-1)*np.sqrt(1-(2* U_E_1-1)**2)+1)/2+0.1*norm.rvs(size=250, loc=0, scale=1)

E_A_12 – means a variable that has been calculated as a sum of two exponentially distributed independent variables:
E_E_1+E_E_2+0.1*norm.rvs(size=250, loc=0, scale=1)

The complete code of the generated variables can be found in the Appendix.

Examples of the introduced dependencies between variables are shown in Figure 5.

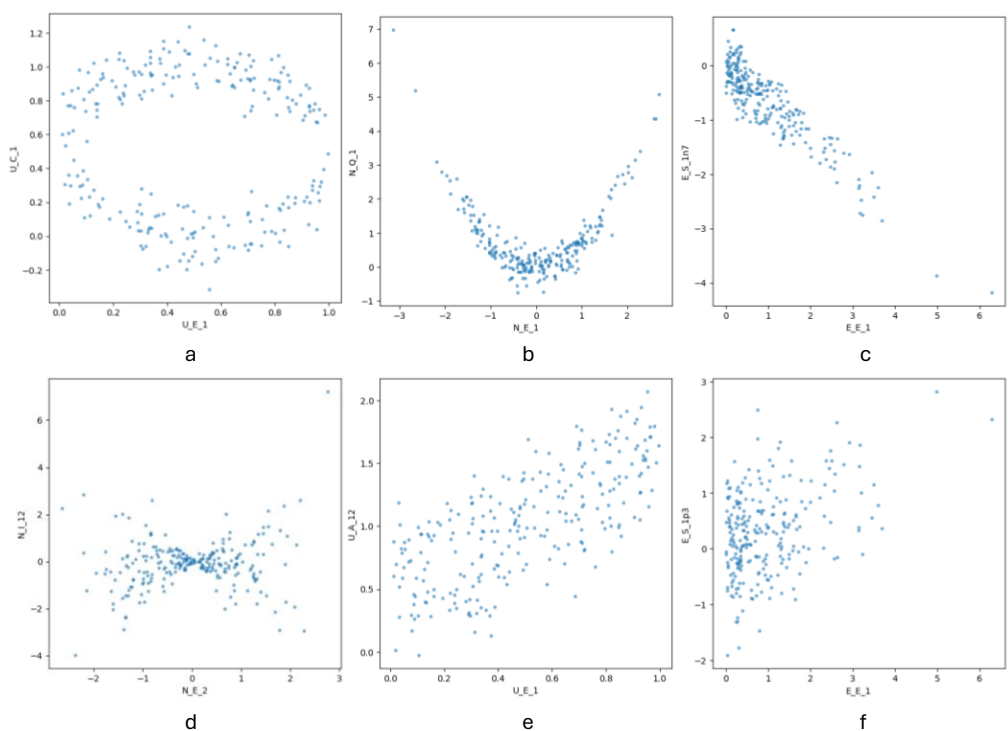

Figure 5. Examples of the introduced dependencies between variables:
(a) circular U_E_1 → U_C_1
(b) quadratic N_E_1² → N_Q_1
(c) negatively correlated linear -E_E_1 → E_S_1n7
(d) interactive N_E_1* N_E_2 → N_I_12
(e) additive U_E_1+ U_E_2 → U_A_12
(f) positively correlated linear E_E_1 → E_S_1p3

## 3.2 Examples of Representations of Variables

Posterior distributions on the 2D latent space ($\mathbb{R}^2$) are forced by β-VAE toward isotropic 2D normal distribution. For better visualization, the latent space has been converted into isotropic uniformed space (-1,+1; -1,+1) by applying the following transformations:

$$Lp_1 = \cos(\alpha)(2\Phi(\rho) - 1)$$
$$Lp_2 = \sin(\alpha)(2\Phi(\rho) - 1)$$

Where $\rho$ and $\alpha$ are radius and angle of the polar coordinate system, consequently, $\rho = \sqrt{Ln_1^2 + Ln_2^2}$, $\cos(\alpha) = Ln_1/\rho$, $\sin(\alpha) = Ln_2/\rho$, $\Phi()$ is CDF of the standard normal distribution, $Ln_{1,2}$ are standardized latent variables $(L_{1,2} - \mu_{1,2})/\sigma_{1,2}$, and $L_{1,2}$ are fitted latent variables.

Figure 6 shows examples of the representations of the synthetic variables on the 2D latent space after applying the above-mentioned transformations.

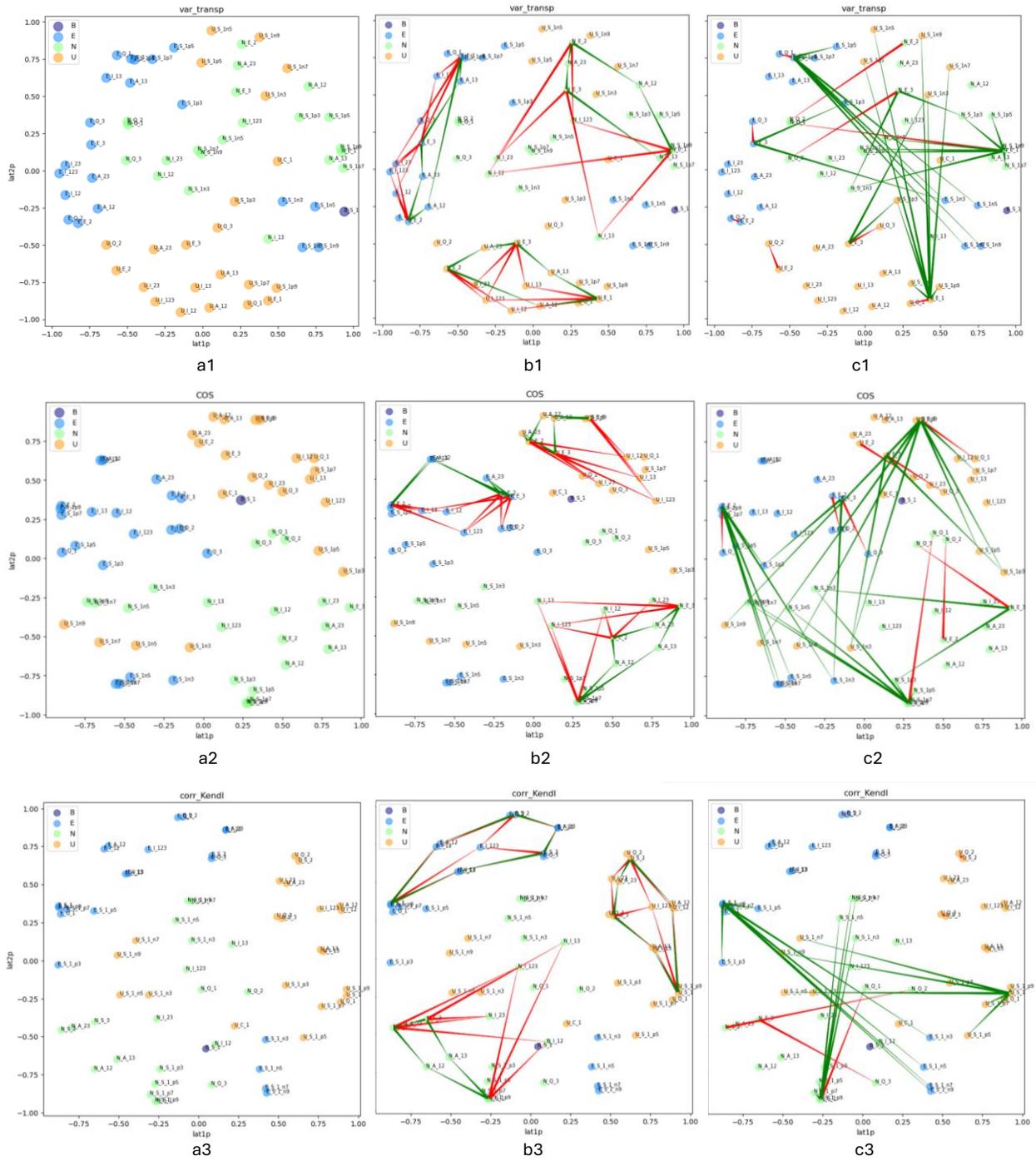

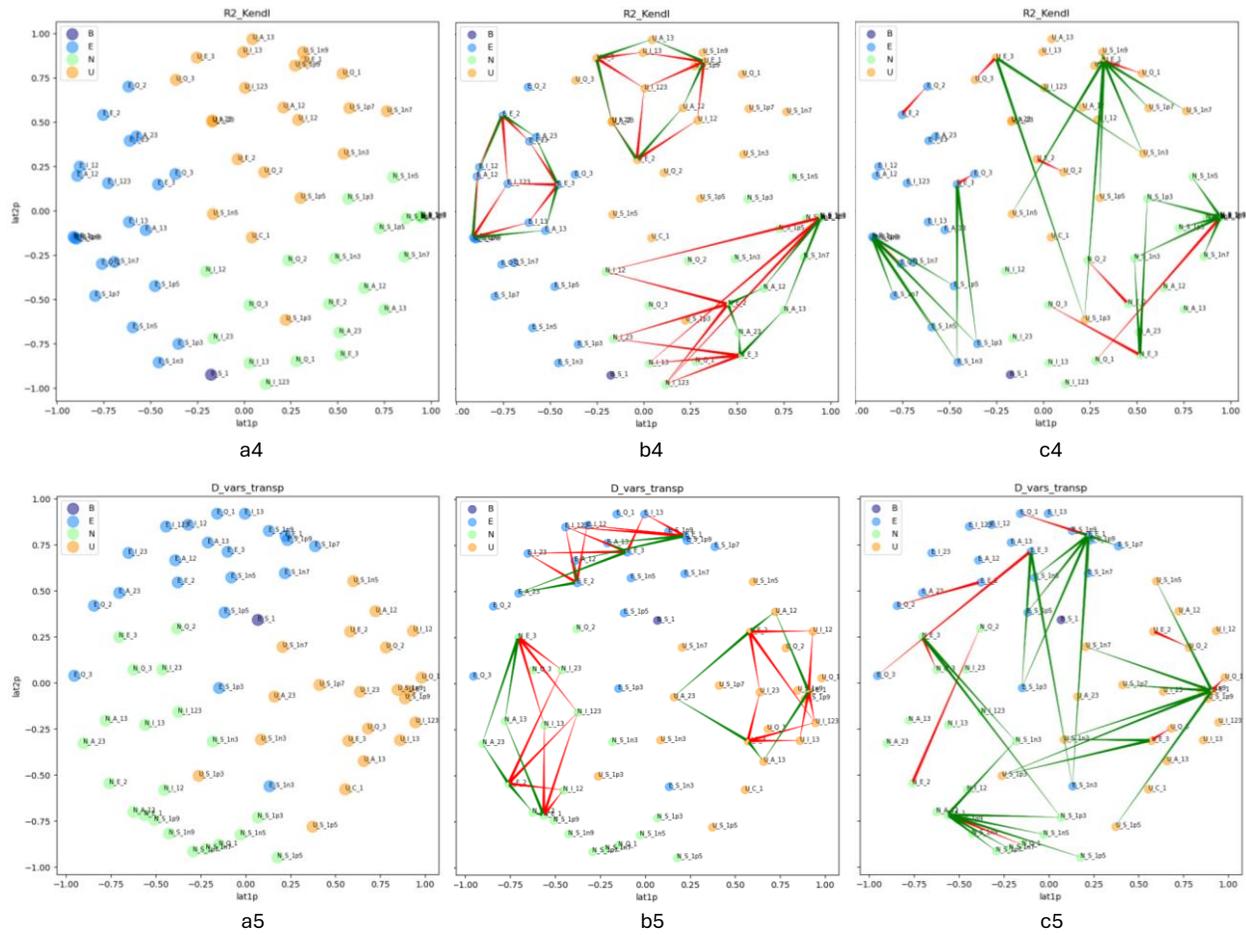

Figure 6. Examples of synthetic variable representations on the 2D latent space:
  (a) scatterplot on the isotropic uniformed space,
  (b) with added green and red lines of additive and interactive dependencies between variables, correspondingly,
  (c) with added green and red lines of linear and non-linear dependencies between variables, correspondingly,
  (1) applying transposed data
  (2) applying pairwise cosine similarity
  (3) applying pairwise Kendall correlation
  (4) applying pairwise Kendall $R^2$
  (5) applying spot magnitudes of the cross product, aggregation option 3
  Thick ends of the lines indicate independent variables

Intuitevely, it can be assumed that pairwise distances between variables on the uniformed latent space will reflect pairwise metrices of the input data. However, it turns out that there are no strong correlations between distances and metrices. It can be attributed to the mapping of the input data by β-VAE encoding to a distribution on the latent space with the following sampling but not to a single point.

Having varying reproducibility of the encoding results by fitting β-VAE, it makes sense to analyse outcomes of the controllable synthetic example by applying qualitative rather than quantitative metrics. Thus, the Quality Function Deployment (QFD) approach (Ginting *et al*, 2020) has been applied to describe results as strong, medium, or weak against selected qualities that are presented in Table 4.

Both Laplacian (D – A) and |J – A| transformations of matrices have been dropped from further consideration. The former has bad performances of representing synthetic variables and their dependencies while the latter has very similar results to the untransformed matrices A.

| Flow | Method | Isotropic Normal | Disentanglement of Distribution Type | Disentanglement of Independent | Alignment of Additives | Alignment of Interactives | Separation between Interactives and Additives | Adjacency of Non-linear Relationships | Adjacency of Positively Linear Relationships | Adjacency of Negatively Linear Relationships |
|---|---|---|---|---|---|---|---|---|---|---|
| A | Transposed dataframe | ● | ○ | ● | ● | ○ | ● | ○ | ● | Δ |
| B | Transposed univariate statistics | ○ | Δ | Δ | Δ | Δ | ● | Δ | Δ | Δ |
|   | Transposed empirical probability density function | ● | ● | ● | ○ | ○ | ● | Δ | ● | ● |
|   | Transposed empirical cumulative density function | ● | ○ | ○ | Δ | Δ | ● | Δ | Δ | Δ |
| C | Pairwise cosine similarity | ● | ○ | ○ | ● | ● | ● | ○ | ○ | Δ |
|   | Pairwise Kendall correlation | ● | ○ | ● | ● | ● | Δ | Δ | ● | Δ |
|   | Pairwise Pearson correlation | ● | ○ | ● | ● | ● | ○ | Δ | ● | Δ |
|   | Pairwise Spearman correlation | ● | ○ | ● | ● | ● | Δ | Δ | ● | Δ |
|   | Pairwise Kendall R-sqr values | ● | ● | ● | ● | ● | Δ | ○ | ● | ● |
|   | Pairwise Pearson R-sqr values | ● | ● | ● | ● | ● | ○ | Δ | ● | ● |
|   | Pairwise Spearman R-sqr values | ● | ● | ● | ● | ● | Δ | ○ | ● | ● |
|   | Pairwise Jaccard index | Δ | Δ | Δ | Δ | Δ | ○ | ○ | ● | ● |
|   | Pairwise Linear Jaccard index | ● | Δ | ● | Δ | Δ | ● | Δ | Δ | Δ |
|   | Pairwise mutual information | ● | ● | ● | ● | ● | ○ | ● | ○ | ○ |
| D | Spot magnitudes of the cross product, aggregation option 1 | Δ | ○ | Δ | Δ | Δ | ● | ○ | ○ | ○ |
|   | Spot magnitudes of the cross product, aggregation option 2 | ● | Δ | ○ | Δ | Δ | Δ | Δ | Δ | Δ |
|   | Spot magnitudes of the cross product, aggregation option 3 | ● | ○ | ● | ○ | ○ | ○ | ○ | ○ | ○ |
|   | Spot magnitudes of the cross product, aggregation option 4 | ● | ● | ● | Δ | Δ | ○ | ○ | ○ | ○ |

Relationships: ● Strong, ○ Medium, Δ Weak

Table 4. Qualitative analysis of representations of synthetic variables on the 2D latent space by different methods

## 4 MNIST Handwritten Digits Example

The second example of the variable representations on the latent space are based on a famous MNIST handwritten 0-9 digits data (LeCun and Cortes, 2010). Illustrations of the handwritten digits are shown in Figure 7, a, that are rendered by 28x28 pixels. This example has been selected for two reasons: first, it includes a large number of 784 pixels, i.e., features, and second, it comes with labels. The latest allow for the observation of both the label and features. Furthermore, the labels that have 0-9 integer values, essentially represent categories by their nature. It allows for consideration of the representation of categorical variables in the example.

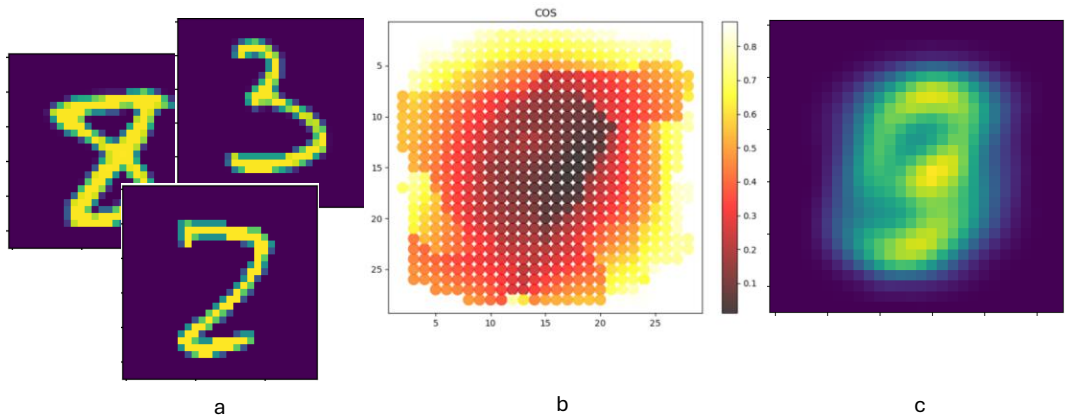

Figure 7. (a) Examples of the MNIST handwritten digits as 28x28 pixel's images; (b) mapping of distances between the label and pixels applying a pairwise cosine similarity metric; and (c) density of the MNIST handwritten lines of 0-9 digits

First, let us consider representation of features only, i.e., pixels without labels (Figure 8, a). The scatterplot chart includes 660 points while 124 pixels have been removed from consideration as having only values equal to zero. It can be observed that there are some patterns but overall, as expected, the representation remains quite isotropic and uniform.

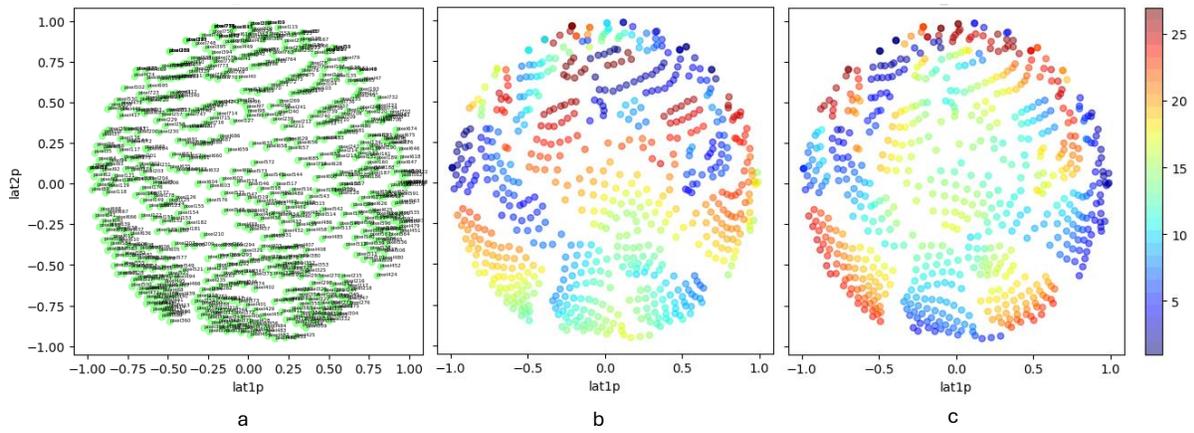

Figure 8. Examples of pixel representation on the isotropic uniformed latent space based on the pairwise matrix of Spearman $R^2$ values without the label:
- (a) with pixel's annotations
- (b) with color mapping of pixel's positions along X-axis of 28x28 images
- (c) with color mapping of pixel's positions along Y-axis of 28x28 images

It turns out that the observed patterns can be attributed to the (X, Y) pixels' position on the 28x28 handwritten images that lead to segmentation and disentanglement of variables (features) on the latent space by unsupervised β-VAE (Figure 8, b and c). Pixel's positions have not been explicitly included in the input of the autoencoder but essential dependencies between pixels of the handwritten images have been emphasized on the fitted latent space representation. The presented example shows that a known disentanglement feature of β-VAE, that has been developed and used to encode observations (Higgins et al, 2017), also successfully sustains the disentanglement abilities while representing variables.

### 4.1 Representations of Labels

Let us explore representations of the MNIST handwritten 0-9 digits but this time with the inclusion of labels. Labels have numeric 0-9 integer values, however, essentially, they represent ten classes and should be recognized as categories. Thus, there is no dependency between the order of integer values in labels and corresponding patterns of pixels. For example, digits 1, 4 and 7 have quite similar handwritten images but their integer values are not sequential. Nevertheless, acknowledging that fact, let us include the labels as a single variable having integer values. Results are shown in Figure 9, a. It can be observed that the label (blue dot) is slightly isolated from the pixel's representations.

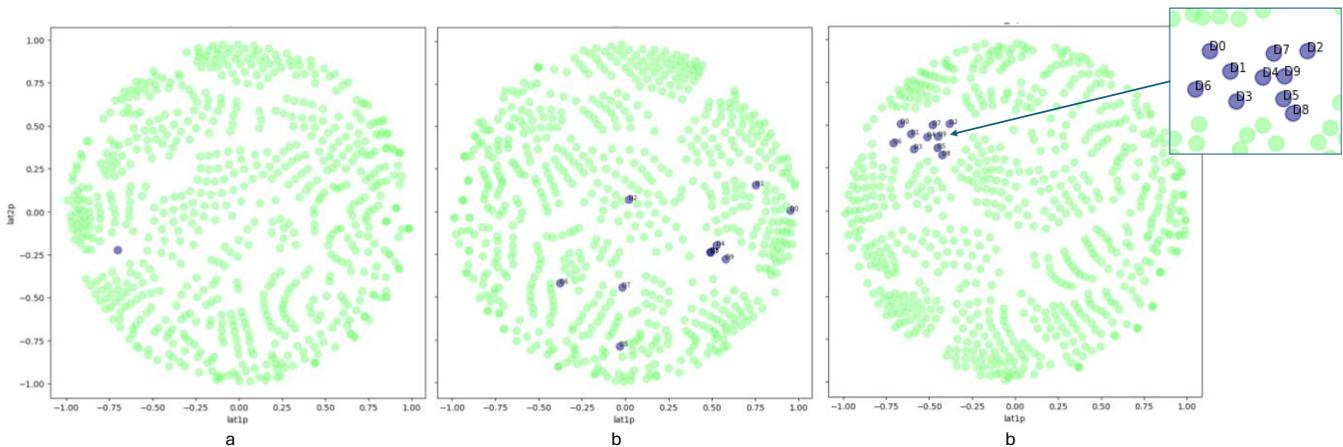

Figure 9. Examples of the representation of pixels (green dots) and labels (blue dots) on the isotropic uniformed latent space based on the pairwise matrix of Spearman $R^2$ values:
- (a) with label represented as an integer
- (b) with one-hot encoded labels
- (c) with reinforced entanglement of one-hot encoded labels

The combined encoding of the pixels and the label allows for calculation of their pairwise distances on the uniform latent space. The latest can be plotted on the 28x28 pixel's space (Figure 7, b). One can observe that central pixels have the shortest distances to the label. On the other hand, this area of images has the highest density of the lines writing 0-9 digits (Figure 7, c).

### 4.2 Reinforced Entanglement of One-Hot Encoded Categorical Variables

To represent categorical variables on the latent space, encoding them as numeric vectors first is required (Jang *et al*, 2017). One-hot encoding is the most popular and straightforward approach that simply transforms the original categorical variable of *M* categories into a vector of *M* dummy binary variables. In our example, *M* equals ten.

After including these ten dummy binary variables, the example of the combined representation of pixels and ten dummy variables on the latent space is shown in Figure 9, b. Even though these ten dummy variables represent encoding of a single categorical variable, they are scattered across the latent space.

To address this issue, the reinforced entanglement of one-hot encoded categories has been introduced. It includes additional *M* columns with the following values: all-one values for the rows corresponding to the one-hot encoded variables, and all-zero values, otherwise (Figure 10).

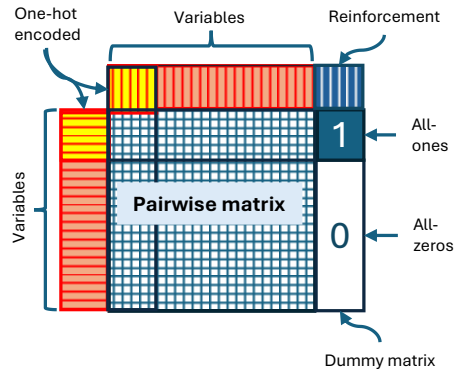

Figure 10. Reinforcement of entanglement of one-hot encoded categorical variable

After applying reinforcement, the dummy variables are tight together on the latent space (Figure 9, c). Furthermore, digits 4, 7, and 9, as well as 5 and 8, that can usually be mutually confused, are placed closer to each other by the autoencoder. The later suggests that inherent pixel dependencies of these digits have been successfully passed through β-VAE with reinforced entanglement of the one-hot encoded labels.

## 5 Canadian Financial Market Example

Let us now consider a real-world example of financial market statistics provided by (Bank of Canada, 2024). It includes twenty-nine time series of Canadian interest rates of eight distinct types and terms having 1,690 weekly observations between 1-Jan-92 and 15-May-24 (Figure 11). In this case, representations of variables on the 2D latent space essentially means representation of multivariate time series.

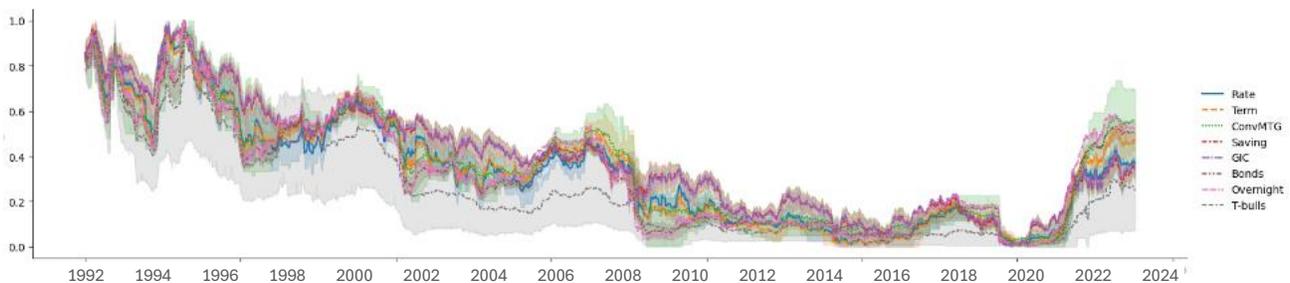

Figure 11. Normalized weekly time series of twenty-nine Canadian interest rates

To apply gradient methods of the rates representations, β-VAE has been trained against observations along time, i.e., weekly records. This is a typical application of variational autoencoders. The original twenty-nine time series of interest rates have been presented as 1,690 points on the 2D latent space (Figure 12, a).

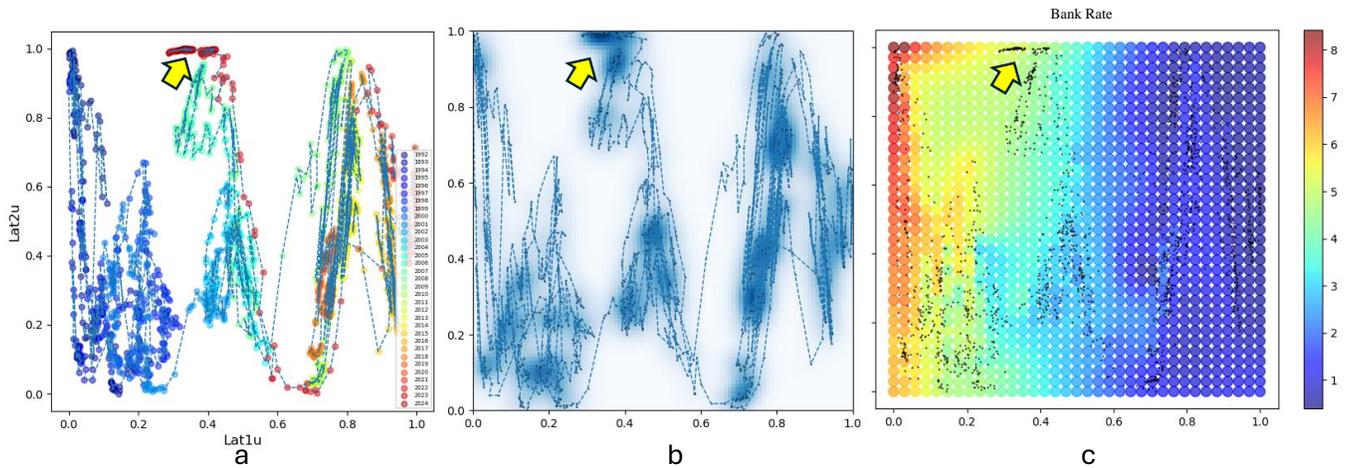

Figure 12. Examples of representations of:
  (a) 1,690 weekly observations between 1-Jan-92 and 15-May-24 of twenty-nine Canadien interest rates time series on the 2D uniformed latent space
  (b) probability densities
  (c) linear interpolation of the Bank rate applying a 35x35 grid
  Yellow arrows point towards the most recent state

There are high-density zones on the 2D uniformed latent space in Figure 12, b despite a Kullback–Leibler divergence term to be part of the applied loss function (Kingma and Welling, 2014; Higgins, 2017). It suggests that there are "attractive" economic states concerning interest rates with some drifts within these zones and quick transitions between high-density zones. It corresponds to stochastically "stable" periods on the time series chart (Figure 11). The unsupervised representation on the latent space allows for isolation of these zones to study underlying conditions.

In addition, it turns out that disengagement on the latent space occurs mostly along a single dimension. According to the example presented in Figure 11, c, values of the Bank interest rate as well as other rates are variated mostly along the X-axis. The latest has a range of 3.0 versus 0.09 of the second latent variable (Y-axis), i.e., about 33 times wider. Considering both the density and the rate mapping charts in Figure 12, b and c, the most recent state (indicated by the yellow arrow) can be seen as a bifurcation point.

Having mapped all twenty-nine interest rates on the observation latent space, it is possible to numerically estimate gradient vectors and then calculate the magnitudes of the cross products as described in sections 2.3 - 2.5 above. The example of the gradient vectors and mapping of the magnitude of the pairwise cross product across the uniformed latent space of observations is presented in Figure 13. Two rates have been selected for the example: the Bank rate and the conventional 3-year term mortgage rate.

The mapping example in Figure 13 reveals that there is a high magnitude zone of the cross product between the Bank and the conventional 3-year term mortgage rates on the upper-left corner of the uniformed latent space. It means that (1) there are high values of the gradients for both variables, i.e., long lengths of vectors, and (2) the orientation of these two vectors are close to orthogonal. High values of the gradients correspond to significant differences along neighboring points on the latent space. Considering the representation of the observations on the latent space (Figure 12, a), the zone of high magnitude values belongs to earlier periods (1992).

2D mapping of the magnitudes of the pairwise cross products in Figure 13 can be presented along time as shown in Figure 14, c.

It turns out that the high values of the magnitudes of the pairwise cross product correspond to periods of high volatility of the underlying interest rates (Figure 14, a and b). Importantly, representation on the observation latent space and, consequently, gradient computations along the X- and Y- axes of the latent space, and the following pairwise cross product calculations were not based on the time variable. Furthermore, input data to fit β-VAE has been randomly reshuffled. Nevertheless, when comparing time series in Figure 14, the calculated magnitudes of the cross products preserve time related variances. It illustrates the potential of the ordering mechanism of β-VAE along the latent variables where time has not been explicitly presented in the training dataset.

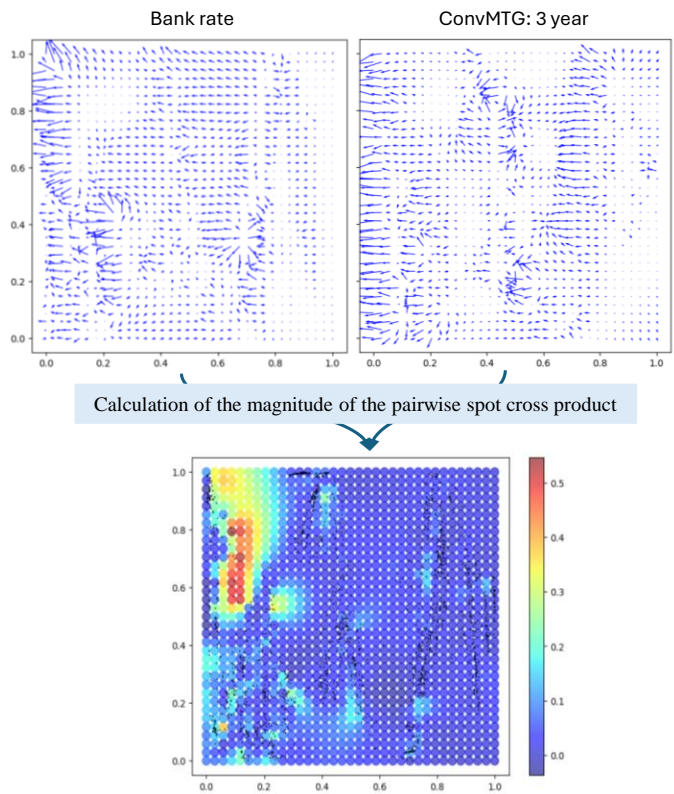

Figure 13. Mapping example of the magnitude of the pairwise cross product across a uniformed latent space of observations

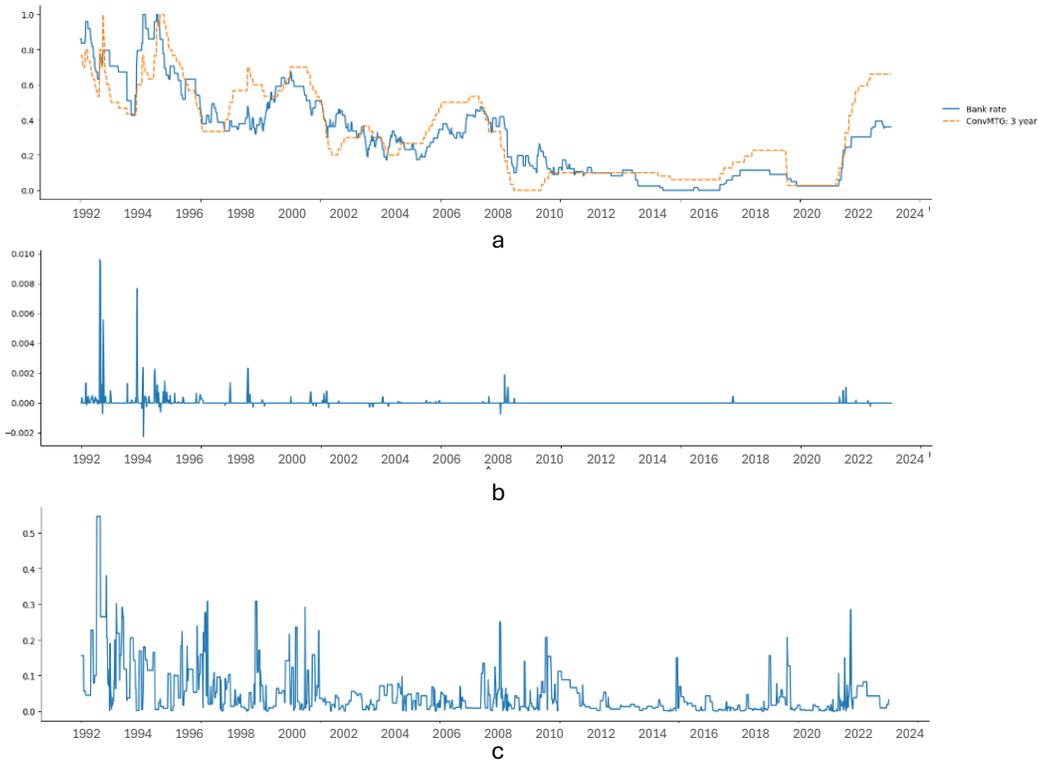

Figure 14. Time series of:
    (a) normalized Bank and conventional 3-year term mortgage rates
    (b) interaction between the first derivatives along time of the selected rates
    (c) magnitude of the cross product between the two selected rates

The following charts provide examples of Canadian interest rates representations on the variable latent space (Figure 15).

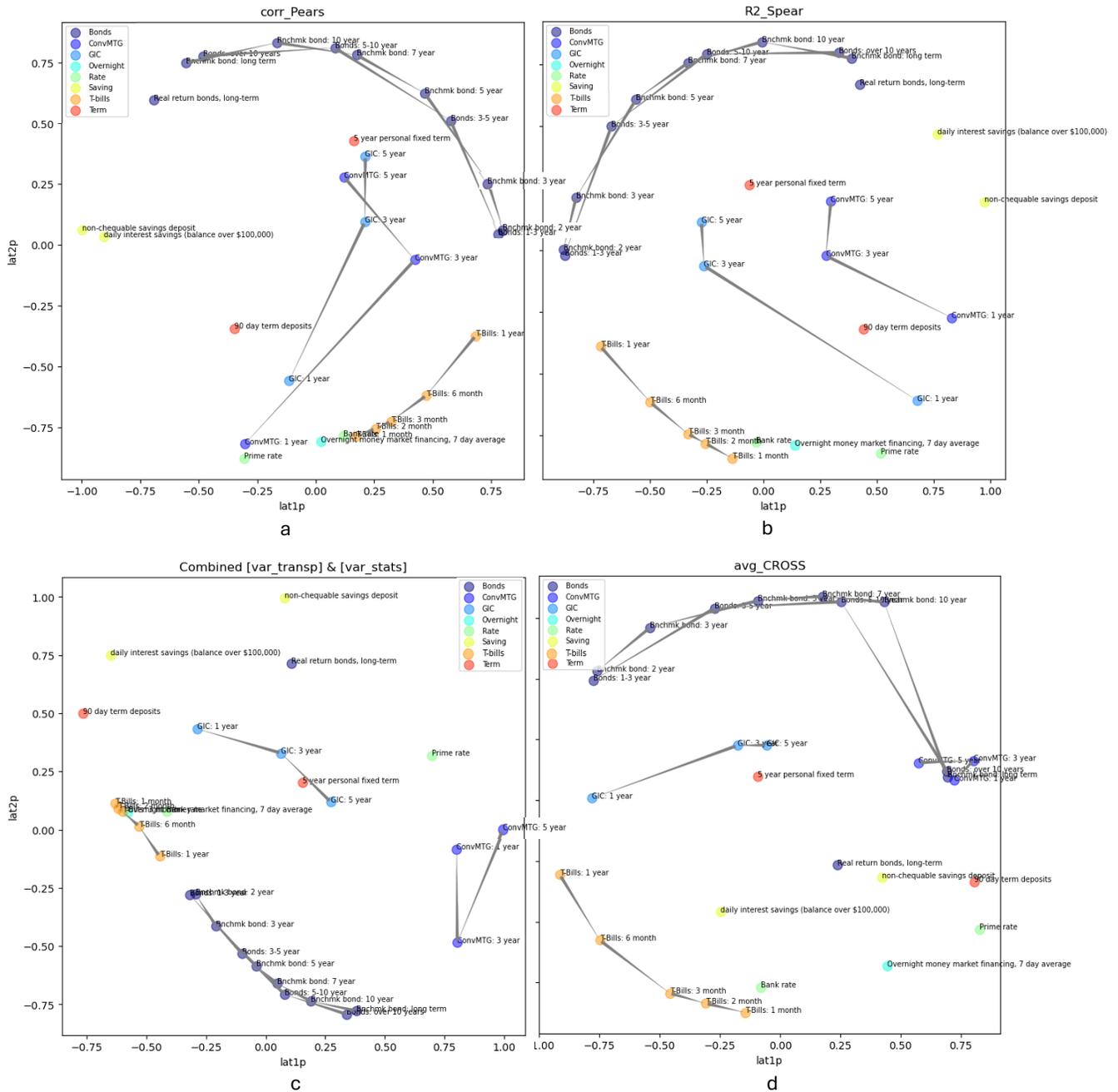

Figure 15. Examples of Canadian interest rates representations on the latent space based on:
   (a) pairwise Pearson correlation matrix
   (b) pairwise Spearman $R^2$ matrix
   (c) combined inputs of transposed data and univariate statistics
   (d) spot magnitudes of the cross product with aggregation option 1
   Color coding corresponds to the type of rates and the thick ends of the pairwise lines indicate rates with a longer term

Figure 15 illustrates dependency of the encoding to the applied methodologies, their underlying metrics, and inherent stochasticity of the β-VAE fitting. Representations based on pairwise Pearson correlation and Spearman $R^2$ matrixes shown in Figure 15, a and b are quite similar. It can be attributed to the fact that the correlation matrix has only positive values and is highly correlated (89%) with a monotonic relationship to the Spearman $R^2$ values.

It can be observed that unsupervised representations of Canadian interest rates on the latent space correctly:
- disentangled rates based on their type, such as bonds, T-bills, GICs, or conventional mortgages

- positioned bonds and T-bills along a single curve
- ordered rates by their terms along the bonds and T-bills curve
- placed Bank and overnight rates closer to the shortest T-Bills of 1 month term

These are quite remarkable results of applying different unsupervised methods. It should be emphasised that the order of rates has been obtained based on inherent data dependencies but not by applying explicit business attributes of the rates such as type or term. It highlights that the representations of variables on the low dimensional latent space can support discovery of important patterns. That observed property is aligned with an initial example of the representation of variables by β-VAE describing electron configurations data in atoms that turned out to exactly match Madelung's rule (Glushkovsky, 2020b).

# 6   Conclusions

The research paper discusses the application of unsupervised variational autoencoders to represent variables on the 2D latent space. Five distinct methods have been introduced: (1) straightforward transposed, (2) univariate metadata of variables, such as variable statistics, empirical PDF and CDF functions, (3) adjacency matrices of different metrics, such as correlations, $R^2$ values, Jaccard index, cosine similarity, or mutual information, (4) gradient mappings of variables on the observation latent space followed by spot cross product calculations, and (5) combined by merging input data from some of the approaches above. Twenty-eight approaches of variable representations by β-VAE have been addressed in the article.

To overcome the challenge of the uniqueness of each variable that corresponds to a single column in the original table, duplications of the input data and random reshuffling followed by deduping of the encoded results has been performed.

Gradient mapping of variables on the observation latent space has been discussed in the paper. By mapping variables on the observation latent space, it is possible to numerically estimate gradient vectors and then to calculate the magnitudes of the cross products. It illustrates the potential of the ordering mechanism of β-VAE along the latent variables.

To estimate gradient vectors for each variable, the following two steps have been performed. First, linear interpolations of values of the variable on the 35x35 grid of the 2D uniformed latent space. Second, numeric estimation of the gradients along the 2D axes. It allows for estimation of gradient vectors of variables for any observation point. Considering vectors of any two variables for the same point on the observation latent space, it allows for the estimation of the pairwise spot magnitude of the cross product between those variables.

Synthetic data has been generated to provide controllable dependencies between variables, such as linear, quadratic, circular, multivariate additive and interactive. This example has been used to assess alternative methodological approaches by observing the obtained representation results against known dependencies.

The article addresses generalized representations of variables that cover both features and labels, as well as continuous numeric and categorical types of data. Dealing with categorical variables, reinforced entanglement has been introduced to represent one-hot encoded categories.

Considering a real-world example of Canadian financial market statistics that includes twenty-nine time series of interest rates of distinct types and terms, it was observed that unsupervised representations of interest rates on the variable latent space correctly: (1) disentangled rates based on their type, such as bonds, T-bills, GICs, or conventional mortgages, (2) positioned bonds and T-bills along a single curve, and (3) ordered rates by their terms along the bonds and T-bills curve. These are quite remarkable results of applying different unsupervised methods. It should be emphasised that the order of rates has been obtained based on inherent data dependencies but not by applying explicit business attributes of the rates such as type or term. It highlights that the representations of variables on the low dimensional latent space can support discovery of important patterns.

## Disclaimer

The paper represents the views of the author and do not necessarily reflect the views of the BMO Financial Group.

**Appendix**

Architecture of the basic β-VAE neural network used in this research to autoencode both observations and variables:

Encoder: Flatten input with *K* channels; Dense (256, activation='relu'); Dense (32, activation='relu'); two Dense (latent_dim=2)

Decoder: Dense (32, input_dim=2, activation='relu'), Dense (512, activation='relu'), Dense (original_dim=*K*)), activation='sigmoid')

Loss=binary_crossentropy+beta*KL; beta=0.3; std=1.0, optimizer=adam; batch_size=128; epochs=75; data sets duplications: train x50 and test x30

Synthetic data frame has been generated by the following code:

```
import pandas as pd
from scipy.stats import norm
from scipy.stats import expon
from scipy.stats import uniform
from scipy.stats import bernoulli

df = pd.DataFrame()
num_obs = 250

df['N_E_1'] = norm.rvs(size=num_obs,loc=0,scale=1)
df['N_E_2'] = norm.rvs(size=num_obs,loc=0,scale=1)
df['N_E_3'] = norm.rvs(size=num_obs,loc=0,scale=1)
df['N_S_1p9'] = 0.9*df.N_E_1+0.1*norm.rvs(size=num_obs,loc=0,scale=1)
df['N_S_1p7'] = 0.7*df.N_E_1+0.3*norm.rvs(size=num_obs,loc=0,scale=1)
df['N_S_1p5'] = 0.5*df.N_E_1+0.5*norm.rvs(size=num_obs,loc=0,scale=1)
df['N_S_1p3'] = 0.3*df.N_E_1+0.7*norm.rvs(size=num_obs,loc=0,scale=1)
df['N_S_1n9'] = -0.9*df.N_E_1+0.1*norm.rvs(size=num_obs,loc=0,scale=1)
df['N_S_1n7'] = -0.7*df.N_E_1+0.3*norm.rvs(size=num_obs,loc=0,scale=1)
df['N_S_1n5'] = -0.5*df.N_E_1+0.5*norm.rvs(size=num_obs,loc=0,scale=1)
df['N_S_1n3'] = -0.3*df.N_E_1+0.7*norm.rvs(size=num_obs,loc=0,scale=1)
df['N_I_12'] = df.N_E_1*df.N_E_2+0.1*norm.rvs(size=num_obs,loc=0,scale=1)
df['N_I_13'] = df.N_E_1*df.N_E_3+0.1*norm.rvs(size=num_obs,loc=0,scale=1)
df['N_I_23'] = df.N_E_2*df.N_E_3+0.1*norm.rvs(size=num_obs,loc=0,scale=1)
df['N_I_123'] = df.N_E_1*df.N_E_2*df.N_E_3+0.1*norm.rvs(size=num_obs,loc=0,scale=1)
df['N_A_12'] = df.N_E_1+df.N_E_2+0.1*norm.rvs(size=num_obs,loc=0,scale=1)
df['N_A_13'] = df.N_E_1+df.N_E_3+0.1*norm.rvs(size=num_obs,loc=0,scale=1)
df['N_A_23'] = df.N_E_2+df.N_E_3+0.1*norm.rvs(size=num_obs,loc=0,scale=1)
df['N_Q_1'] = 0.7*df.N_E_1**2+0.3*norm.rvs(size=num_obs,loc=0,scale=1)
df['N_Q_2'] = 0.7*df.N_E_2**2+0.3*norm.rvs(size=num_obs,loc=0,scale=1)
df['N_Q_3'] = 0.7*df.N_E_3**2+0.3*norm.rvs(size=num_obs,loc=0,scale=1)
df['U_E_1'] = uniform.rvs(size=num_obs,loc=0,scale=1)
df['U_E_2'] = uniform.rvs(size=num_obs,loc=0,scale=1)
df['U_E_3'] = uniform.rvs(size=num_obs,loc=0,scale=1)
df['U_S_1p9'] = 0.9*df.U_E_1+0.1*norm.rvs(size=num_obs,loc=0,scale=1)
df['U_S_1p7'] = 0.7*df.U_E_1+0.3*norm.rvs(size=num_obs,loc=0,scale=1)
df['U_S_1p5'] = 0.5*df.U_E_1+0.5*norm.rvs(size=num_obs,loc=0,scale=1)
df['U_S_1p3'] = 0.3*df.U_E_1+0.7*norm.rvs(size=num_obs,loc=0,scale=1)
df['U_S_1n9'] = -0.9*df.U_E_1+0.1*norm.rvs(size=num_obs,loc=0,scale=1)
df['U_S_1n7'] = -0.7*df.U_E_1+0.3*norm.rvs(size=num_obs,loc=0,scale=1)
df['U_S_1n5'] = -0.5*df.U_E_1+0.5*norm.rvs(size=num_obs,loc=0,scale=1)
df['U_S_1n3'] = -0.3*df.U_E_1+0.7*norm.rvs(size=num_obs,loc=0,scale=1)
df['U_I_12'] = df.U_E_1*df.U_E_2+0.1*norm.rvs(size=num_obs,loc=0,scale=1)
df['U_I_13'] = df.U_E_1*df.U_E_3+0.1*norm.rvs(size=num_obs,loc=0,scale=1)
df['U_I_23'] = df.U_E_2*df.U_E_3+0.1*norm.rvs(size=num_obs,loc=0,scale=1)
df['U_A_12'] = df.U_E_1+df.U_E_2+0.1*norm.rvs(size=num_obs,loc=0,scale=1)
df['U_A_13'] = df.U_E_1+df.U_E_3+0.1*norm.rvs(size=num_obs,loc=0,scale=1)
df['U_A_23'] = df.U_E_2+df.U_E_3+0.1*norm.rvs(size=num_obs,loc=0,scale=1)
df['U_Q_1'] = 0.7*df.U_S_1**2+0.1*norm.rvs(size=num_obs,loc=0,scale=1)
df['U_Q_2'] = 0.7*df.U_E_2**2+0.1*norm.rvs(size=num_obs,loc=0,scale=1)
df['U_Q_3'] = 0.7*df.U_E_3**2+0.1*norm.rvs(size=num_obs,loc=0,scale=1)
df['U_I_123'] = df.U_E_1*df.U_E_2*df.U_E_3+0.1*norm.rvs(size=num_obs,loc=0,scale=1)
df['U_C_1'] = ((2*bernoulli.rvs(size=num_obs, p=0.5)-1)*np.sqrt(1-(2*df.U_E_1-1)**2)+1)/2+0.1*norm.rvs(size=num_obs,loc=0,scale=1)
df['E_E_1'] = expon.rvs(size=num_obs,loc=0,scale=1)
df['E_E_2'] = expon.rvs(size=num_obs,loc=0,scale=1)
df['E_E_3'] = expon.rvs(size=num_obs,loc=0,scale=1)
df['E_S_1p9'] = 0.9*df.E_E_1+0.1*norm.rvs(size=num_obs,loc=0,scale=1)
df['E_S_1p7'] = 0.7*df.E_E_1+0.3*norm.rvs(size=num_obs,loc=0,scale=1)
df['E_S_1p5'] = 0.5*df.E_E_1+0.5*norm.rvs(size=num_obs,loc=0,scale=1)
```

```python
df['E_S_1p3'] = 0.3*df.E_E_1+0.7*norm.rvs(size=num_obs,loc=0,scale=1)
df['E_S_1n9'] = -0.9*df.E_E_1+0.1*norm.rvs(size=num_obs,loc=0,scale=1)
df['E_S_1n7'] = -0.7*df.E_E_1+0.3*norm.rvs(size=num_obs,loc=0,scale=1)
df['E_S_1n5'] = -0.5*df.E_E_1+0.5*norm.rvs(size=num_obs,loc=0,scale=1)
df['E_S_1n3'] = -0.3*df.E_E_1+0.7*norm.rvs(size=num_obs,loc=0,scale=1)
df['E_I_12'] = df.E_E_1*df.E_E_2+0.1*norm.rvs(size=num_obs,loc=0,scale=1)
df['E_I_13'] = df.E_E_1*df.E_E_3+0.1*norm.rvs(size=num_obs,loc=0,scale=1)
df['E_I_23'] = df.E_E_2*df.E_E_3+0.1*norm.rvs(size=num_obs,loc=0,scale=1)
df['E_A_12'] = df.E_E_1+df.E_E_2+0.1*norm.rvs(size=num_obs,loc=0,scale=1)
df['E_A_13'] = df.E_E_1+df.E_E_3+0.1*norm.rvs(size=num_obs,loc=0,scale=1)
df['E_A_23'] = df.E_E_2+df.E_E_3+0.1*norm.rvs(size=num_obs,loc=0,scale=1)
df['E_Q_1'] = 0.7*df.E_E_1**2+0.3*norm.rvs(size=num_obs,loc=0,scale=1)
df['E_Q_2'] = 0.7*df.E_E_2**2+0.3*norm.rvs(size=num_obs,loc=0,scale=1)
df['E_Q_3'] = 0.7*df.E_E_3**2+0.3*norm.rvs(size=num_obs,loc=0,scale=1)
df['E_I_123'] = df.E_E_1*df.E_E_2*df.E_E_3+0.1*norm.rvs(size=num_obs,loc=0,scale=1)
df['B_E_1'] = bernoulli.rvs(size=num_obs, p=0.5)
```